\documentclass[lettersize,journal]{IEEEtran}
\usepackage{amsmath,amsfonts}
\usepackage{algorithmic}
\usepackage{algorithm}
\usepackage{hyperref}
\usepackage{array}
\usepackage{textcomp}
\usepackage{stfloats}
\usepackage{url}
\usepackage{verbatim}
\usepackage{graphicx}
\usepackage{booktabs}
\usepackage{amsthm}
\newtheorem{theorem}{Theorem}[section]

\newtheorem{proposition}[theorem]{Proposition}

\usepackage{array}
\usepackage{xcolor}
\usepackage{cite}
\usepackage{multirow}
\usepackage{bm}

\usepackage{pgfplots}
\usepackage{tikz}
\usetikzlibrary{positioning}
\usepackage{amsthm}
\usepackage{amsmath, amsthm, amssymb}

\definecolor{peach}{RGB}{255, 228, 196} 
\definecolor{lightpeach}{RGB}{255, 240, 210}

\definecolor{c11}{rgb}{0.2, 0.4, 0.6}  
\definecolor{c12}{rgb}{0.6, 0.4, 0.2}  
\definecolor{c13}{rgb}{0.4, 0.6, 0.2}  
\definecolor{c14}{rgb}{0.5, 0.3, 0.5}  
\definecolor{c15}{rgb}{0.6, 0.2, 0.2}  

\definecolor{c21}{rgb}{0.2, 0.4, 0.6}  
\definecolor{c22}{rgb}{0.6, 0.4, 0.2}  
\definecolor{c23}{rgb}{0.4, 0.6, 0.2}  
\definecolor{c24}{rgb}{0.5, 0.3, 0.5}  
\definecolor{c25}{rgb}{0.6, 0.2, 0.2}  

\definecolor{c31}{rgb}{0.0, 0.6, 1.0}
\definecolor{c32}{rgb}{0.1, 0.4, 0.7}
\definecolor{c33}{rgb}{0.2, 0.2, 0.4}

\begin{document}

\title{
Deep Learning Optimization of Two-State Pinching Antennas Systems
}

\author{Odysseas G. Karagiannidis, ~\IEEEmembership{Student Member,~IEEE}, Victoria E. Galanopoulou, ~\IEEEmembership{Student Member,~IEEE}, Panagiotis D. Diamantoulakis,~\IEEEmembership{Senior Member,~IEEE}, Zhiguo Ding, ~\IEEEmembership{Fellow,~IEEE}, Octavia Dobre,~\IEEEmembership{Fellow,~IEEE}
\thanks{O. G. Karagiannidis, V. E. Galanopoulou, and P. D. Diamantoulakis are with the Aristotle University of Thessaloniki, 54124 Thessaloniki, Greece (e-mails: \{okaragia, vgalanop, padiaman\}@ece.auth.gr). Z. Ding is with the School of Electrical and Electronic Engineering, The University of Manchester, Manchester, M13 9PL, UK (e-mail: zhiguo.ding@ieee.org). O. Dobre is with the Faculty of Engineering and Applied Science, Memorial University, Canada (e-mail: odobre@mun.ca).}}




\maketitle

\begin{abstract}
The evolution of wireless communication systems requires flexible, energy-efficient, and cost-effective antenna technologies. Pinching antennas (PAs), which can dynamically control electromagnetic wave propagation through binary activation states, have recently emerged as a promising candidate. In this work, we investigate the problem of optimally selecting a subset of fixed-position PAs to activate in a waveguide, when the aim is to maximize the communication rate at a user terminal. Due to the complex interplay between antenna activation, waveguide-induced phase shifts, and power division, this problem is formulated as a combinatorial fractional 0-1 quadratic program. To efficiently solve this challenging problem, we use neural network architectures of varying complexity to learn activation policies directly from data, leveraging spatial features and signal structure. Furthermore, we incorporate user location uncertainty into our training and evaluation pipeline to simulate realistic deployment conditions. Simulation results demonstrate the effectiveness and robustness of the proposed models.
\end{abstract}


\vspace{-4mm}
\section{Introduction}

The future of wireless communication technology requires antennas that can operate efficiently, affordably, and flexibly in dynamic settings.
Traditional static antennas with fixed geometry and hardware-dependent beamforming often lack the adaptability required in dynamic wireless environments. Consequently, there is a growing interest in flexible antenna systems that can dynamically reconfigure wireless channels to enhance reliability, reduce path loss, and increase spectral efficiency.
Promising solutions among such technologies include reconfigurable intelligent surfaces (RISs), fluid antennas, and movable antennas. RISs use reconfigurable surfaces to redirect incoming signals by adjusting electromagnetic parameters, which offers enhanced coverage in non-line-of-sight (NLoS) conditions. However, RISs often experience double attenuation due to signal reflection, which limits their practical deployment. Meanwhile, fluid and movable antennas introduce physical mobility to adapt antenna positions dynamically for optimized performance \cite{ref1} \cite{ref2}, \cite{ref3}, \cite{ref4}, \cite{ref5}. However, their adjustment range is usually limited to a few wavelengths, which restricts their effectiveness against large-scale path loss and severe channel fading. Recent developments also explore the integration of generative artificial intelligence (AI) and multi-agent large language models (LLMs) into 6G system design and simulation frameworks to enable intelligent automation and adaptability \cite{AI_6G, llm_6g}. These advancements suggest that, beyond physical antenna flexibility, software-driven intelligence may play a critical role in managing the next-generation wireless infrastructure under highly dynamic conditions.
In light of these limitations, pinching antennas (PAs), introduced by DOCOMO, have emerged as a new and promising direction within the landscape of flexible-antenna technologies \cite{docomo2021pinching}.
Unlike fluid or movable antennas, which typically allow for small-scale position shifts, PAs leverage dielectric particles along a waveguide to dynamically activate specific radiation points.  As the electromagnetic signals propagating along the waveguide will leak from the multiple antennas, PAs share certain characteristics with leaky-wave antennas \cite{ref6}, which have been recently explored for applications such as holographic multiple-input multiple-output \cite{leaky_mimo}. However, compared to leaky-wave-based systems where antenna spacing remains constrained to the wavelength scale, limiting their ability to mitigate large-scale path loss, PAs enable dynamic activation at subwavelength intervals along a dielectric waveguide.
This spatial flexibility of PAs enables robust performance in obstructed environments, effectively mitigating path loss and establishing stable line-of-sight (LoS) links, even in challenging scenarios \cite{ref7}.  Recent studies show that, the performance gain of pinching antenna systems (PASs) compared to conventional antennas increases with the increase the LoS blockage probability \cite{los}.  In parallel, recent work has also examined the theoretical array gain achievable in PA architectures, providing fundamental insights into how constructive signal superposition can be leveraged  under practical constraints \cite{array_gain}. With low fabrication complexity and minimal energy demands, PAs offer a scalable and cost-effective approach to overcoming coverage challenges in dynamic environments \cite{yang2025pinching}.
\subsection{State-of the-Art}
Existing works on PASs primarily investigate optimization strategies based on the spatial placement of antennas to improve coverage or maximize system throughput in both downlink and uplink scenarios  \cite {zeng2025resource}, \cite{ref8},\cite{ref9},\cite{nallanathan_uplink}. In \cite{tyrovolas}, the authors presented closed-form expressions for the outage probability and ergodic rate of pinching-antenna systems (PASs), highlighting that waveguide losses, especially in extended configurations, can significantly degrade system performance. Recent studies have also explored the security capabilities of PASs, introducing artificial noise–aided beamforming and optimized PA positioning to enhance secrecy rates and improve resilience against eavesdropping attacks in both single- and multi-waveguide configurations \cite{papanikolaou},\cite{liu_security}. Recent efforts have also advanced the application of PASs in integrated sensing and communication (ISAC), where joint designs for sensing and data transmission leverage the spatial flexibility of PAs to improve resolution and system efficiency \cite{isac1, isac2}. More recently, learning-based methods such as graph neural networks (GNNs) have been explored to solve the antenna scheduling problem in complex scenarios \cite{ref10}. However, most of these studies model the position as a continuous optimization variable, which may not be realistic in practice. In realistic deployments, PAs are typically installed at discrete preconfigured positions, and only their activation states (i.e., opened or closed) can be dynamically controlled. This practical formulation simplifies the system design, mitigates the delays associated with mechanical repositioning, and transforms the complex antenna placement problem into a more tractable activation scheduling task \cite{ref11}, \cite{xieplacement}. Building on this aspect, recent advancements have explored antenna activation strategies for PASs with non-orthogonal multiple access (NOMA) with fixed antenna positions, aiming to optimize system throughput  \cite{wang2025antenna}, \cite{zeng2025energy}.
\subsection{Contribution}
In this work, we adopt an activation-based scheme and focus on selecting the optimal subset of antennas to activate from a statically deployed pinching-antenna system in order to maximize the user's communication rate. s previously mentioned, the proposed scheme is more practical than approaches that treat the PA positions as continuous variables. Moreover, it can also be applied to such cases by discretizing the position space, i.e.,  by selecting a subset of a much larger set of potential PA locations, with the subset size equal to the number of available PAs.

The complex interaction of electromagnetic phases across antennas and the nonlinear effect of power division among active elements on the achievable rate complicates this problem, even under static geometry. This gives rise to a challenging antenna activation problem that we formulate as a quadratic fractional 0-1 programming problem, a known class of nonconvex combinatorial problems.

Neural networks are employed to efficiently solve the antenna activation scheduling problem due to their ability to approximate complex relationships from data and enable real-time inference. Unlike conventional optimization approaches, which are computationally intensive and often require full system knowledge, neural models offer a scalable, data-driven alternative. As a starting point, we consider a multilayer perceptron (MLP), which predicts antenna activations based on signal strength and phase information. While this model captures basic patterns, it lacks spatial understanding. To improve upon this model, we use a GNN and attention mechanisms to model the antenna-user system as a graph, which enables spatial reasoning. GNNs are well-suited to this task because they naturally capture the spatial and connectivity patterns of wireless systems and generalize effectively across different configurations \cite{gnn_drl},\cite{GNN_MIMO},\cite{gnn_nallanathan}. 

Finally, the proposed scheme is extended and evaluated under position uncertainty, which is an inherent challenge in PASs arising from location estimation errors, affecting the constructive superposition of delayed signal components at the receiver. To enhance the model's robustness, we introduce a modified activation policy that accounts for this uncertainty and incorporate it into the simulations. This approach enables reliable antenna activation even in the presence of imprecise location estimates.

 \subsection{Structure}
The paper is organized as follows. In Section II, we describe the system model, including the waveguide geometry and channel model. In Section III,  the antenna activation problem is formulated as a fractional quadratic 0-1 program. In Section IV, we introduce the proposed deep learning-based approach and describe three architectures of increasing complexity. Section V extends the existing framework adding uncertainty in user positions. Section VI presents our experimental results and compares model performance across various scenarios. Finally, Section VII concludes the paper.

\section{System Model}

According to Fig. \eqref{fig:system_model}, we consider a single waveguide located at height \( H \), aligned along the \( x \)-axis and centered at position \( y = 0\).
Let the length of the waveguide be \( D \), and \( N \) denote the number of uniformly spaced PAs placed along the waveguide. Then, the location of the \( n \)-th PA is given by:

\begin{figure}[t!]
    \centering
    \includegraphics[width=1.15\linewidth]{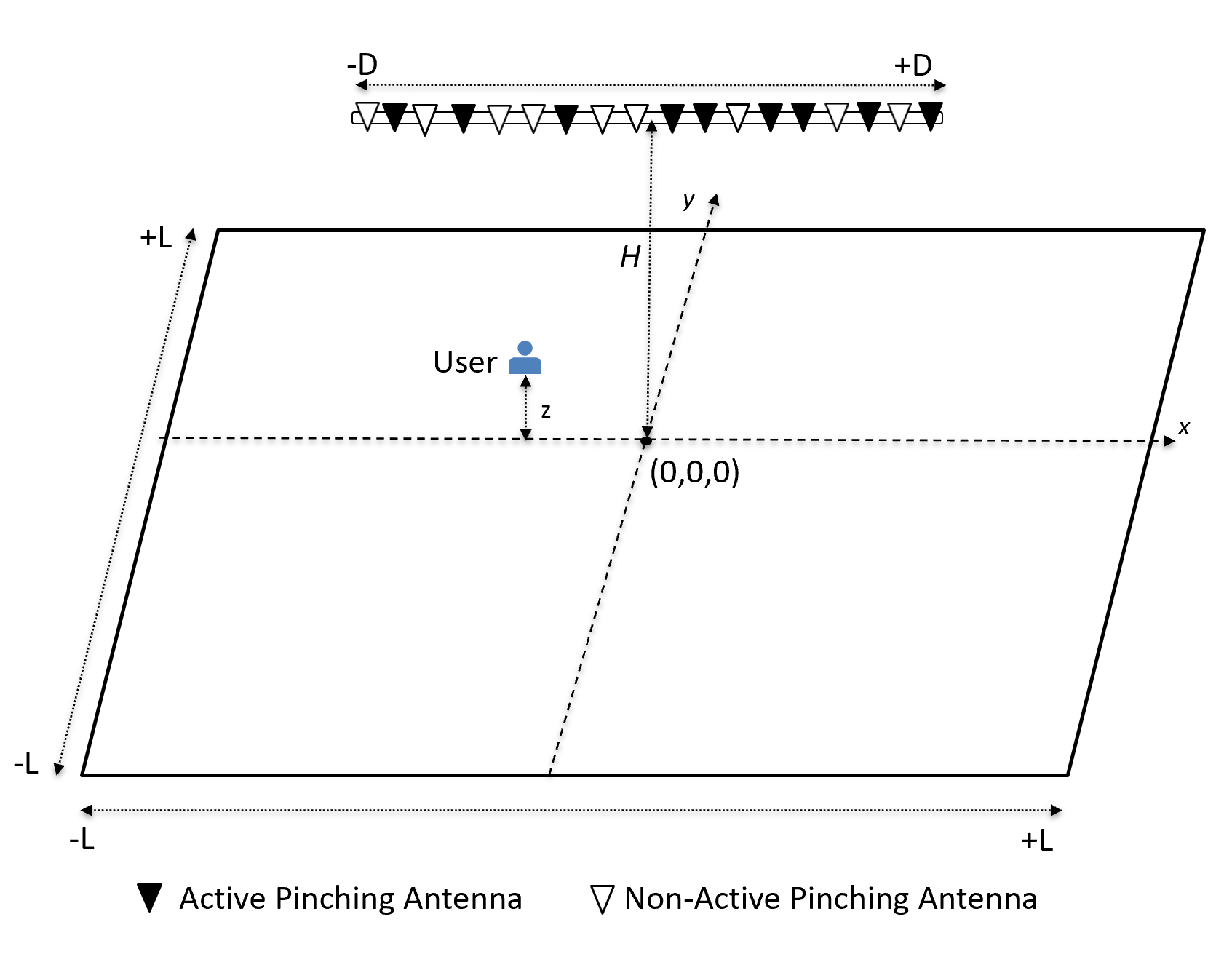}
    \caption{Pinching antenna system}
    \label{fig:system_model}
\end{figure}
\begin{equation}
    \bm{\psi}^{\mathrm{Pin}}_n = \left( -D +\frac{2D}{N -1 }n,\ 0 ,\ H \right), \quad n =  0, \dots, N-1.
\end{equation}

The set of all antenna positions can be represented by the vector:
\begin{equation}
    \bm{\psi}^{\text{Pin}} = 
    \begin{bmatrix}
        \bm{\psi}^{\text{Pin}}_0 \\
        \bm{\psi}^{\text{Pin}}_1 \\
        \vdots \\
        \bm{\psi}^{\text{Pin}}_{N-1}
    \end{bmatrix}.
\end{equation}

We assume a time division multiple access (TDMA) protocol, where only one user is scheduled at a time for communication. This approach allows the system to concentrate all available transmission power and beamforming resources on a single terminal during each time slot, thereby simplifying the analysis and maximizing individual user performance.
The user terminals are deployed on the ground plane at height $z$, uniformly within a square region of side length \( D \). That is, the location of the considered user is:
\begin{equation}
    \bm{\psi} = (x,\ y,\ z),
    \quad \text{with }
    x,\ y \in [-L , L ] ,\ z \in [0,1].
\end{equation}

In later sections, we consider both perfect and imperfect knowledge of the user’s position. In particular, for the case of imperfect position knowledge, we examine the impact of realistic position estimation accuracy by accounting for estimation uncertainties on the ground plane and modeling the user’s position as a Gaussian random variable.

\subsection{Received Signal Model}

We assume that only a subset of the \( N \) PAs is activated. Let \( a_n \in \{0, 1\} \) be the binary activation variable for antenna \( n \), and define the beam-forming vector:
\begin{equation}
\mathbf{a} = [a_0,\ a_1,\ \dots,\ a_{N-1}]^T,
\end{equation}
where $[\cdot]^T$ is the transpose of the vector $[\cdot]$.  Let \( s \in \mathbb{C} \) denote the transmitted signal (with power \( P \)), and 
 the channel gain from the \(n\)-th PA to user be:
\begin{equation}
    h_n =  \frac{\exp\left(-j \frac{2\pi}{\lambda} \left\| \bm{\psi} - \bm{\psi}^{\text{Pin}}_n \right\| \right)}{\left\| \bm{\psi} - \bm{\psi}^{\text{Pin}}_n \right\|}, \quad n = 0, 1, \dots, N-1,
\end{equation}
The received signal at the user is
\begin{equation}
    y = \sqrt{\frac{P\eta }{N_{a}}}  \sum_{n=1}^{N} a_n h_ng_ns + w,
\end{equation}
where \(N_{a}\) is the number of activated PAs, \( s \in \mathbb{C} \) denotes the transmitted signal (with power \( P \)), \( \eta \) accounts for the path loss scaling factor, $h_n$ is the channel gain from the \(n\)-th PA defined as
\begin{equation}
    h_n =  \frac{\exp\left(-j \frac{2\pi}{\lambda} \left\| \bm{\psi} - \bm{\psi}^{\text{Pin}}_n \right\| \right)}{\left\| \bm{\psi}_m - \bm{\psi}^{\text{Pin}}_n \right\|}, \quad n = 0, 1, \dots, N-1,
\end{equation}
\(w \sim \mathcal{CN}(0, \sigma^2_w)\) is the additive white Gaussian noise, and
\begin{equation}
g_n=e^{-j\theta_n}
\end{equation}
is the channel in the waveguide, with 
\begin{equation}
\theta_n=\dfrac{2\pi}{\lambda_g} \left\| \bm{\psi}_n^{\text{Pin}} - \bm{\psi}_0^{\text{Pin}} \right\|
\end{equation}
being  the waveguide-induced phase shift from the feed point \(\bm{\psi}_0^{\text{Pin}}\) to the \(n\)-th PA.


\subsection{Achievable Rate}

The received signal power at the user, accounting for the selected antennas and the waveguide-induced phase shifts, is given by:
\begin{equation}
    P_{\text{r}} = \frac{P\eta}{N_{a}} \left| \sum_{n=1}^N a_n h_{n} e^{-j \theta_n} \right|^2,
\end{equation}
where $N_a=\left\lVert a \right\rVert_0$ is the number of activated PAs.

Accordingly, the received signal-to-noise ratio (SNR) becomes:
\begin{equation}
    \gamma = \frac{P\eta}{N_{a} \sigma_w^2} \left| \sum_{n=1}^N a_n h_{n,m} e^{-j \theta_n} \right|^2,
\end{equation}
and the achievable communication rate of the user $u_m$ is:
\begin{equation}
    R = \log_2 \left( 1 + \gamma_m \right).
\end{equation}

To facilitate a more compact representation, we define the complex effective gain of the $n$-th antenna as
\begin{equation}
    B_n = h_n e^{-j \theta_n}, \quad \text{for } n = 0, \dots, N-1,
\end{equation}
and collect these into the vector
\begin{equation}
    \mathbf{B} = [B_0, B_1, \dots, B_{N-1}]^T \in \mathbb{C}^{N}.
\end{equation}

Let $\mathbf{a} \in \{0, 1\}^N$ be the binary selection vector indicating the activation state of each PA. Then, the received signal at the user can be written as
\begin{equation}
    y_m = \sqrt{\frac{P\eta}{N_{a}}} \mathbf{a}^T \mathbf{B}  s + w.
\end{equation}
Accordingly, the received SNR can also be expressed in terms of  \( \mathbf{B} \) as:
\begin{equation}
    \gamma_m = \frac{P \eta}{N_{\text{a}} \sigma_w^2} \left| \mathbf{a}^T \mathbf{B} \right|^2.
    \label{eq:snr_bform}
\end{equation}

\vspace{1em}
\section{Optimization Problem }

In the considered PAS, the total transmit power $P$ is equally divided among the active antennas. As a result, activating a larger number of antennas distributes the power more thinly across the set, reducing the power available per antenna. While multiple antennas can combine their signals constructively to enhance the received power, this benefit must be balanced against the power splitting penalty.
Moreover, each PA contributes a signal with a distinct phase, and the optimization must account for phase alignment in order to ensure coherent signal combining at the receiver.

We seek to select the optimal binary activation vector $\mathbf{a} \in \{0,1\}^N$ that maximizes the user's rate ($R_m(\mathbf{a})$) and as a result the received signal power. This leads to the following non-linear combinatorial optimization problem:

\begin{equation}
    \max_{\mathbf{a} \in \{0,1\}^N} \frac{|\mathbf{a}^T \mathbf{B}|^2}{\left\lVert \mathbf{a} \right\rVert_0}.
        \label{eq:max_snr} 
\end{equation} \label{eq:opt_problem}

\subsection{Quadratic Fractional 0-1 Formulation}

We model our antenna selection task as a special case of a quadratic fractional 0-1 programming (QF01P) problem, a well-known class of nonconvex combinatorial optimization. 

\begin{proposition}
Assuming perfect knowledge of the user's location, the sum rate maximization problem  can be written a special case of a QF01P problem, a well-known class of nonconvex combinatorial optimization, and be expressed as
\begin{equation}\label{eq:our_qf01p} 
\begin{array}{cl}
\underset{\mathbf{x}}{\bm{\max}} & \displaystyle\frac{\mathbf{a}^\top \mathbf{Q} \mathbf{a}}{\mathbf{1}^\top \mathbf{a}} \\\\
\textbf{s.t.} \quad  &\mathrm{C}_1:\quad \mathbf{a} \in \{0,1\}^n 
\end{array}
\end{equation}
with \begin{equation}
\mathbf{Q} = \Re(\mathbf{B} \mathbf{B}^H).
\end{equation}
\end{proposition}

\begin{IEEEproof}

The general form of a QF01P is given by:

\begin{equation}\label{eq:1stopt} 
\begin{array}{cl}
\underset{\mathbf{x}}{\bm{\max}} & \displaystyle\frac{\mathbf{x}^\top \mathbf{Q} \mathbf{x} + \mathbf{c}^\top \mathbf{x} + r}{\mathbf{x}^\top D \mathbf{x} + \mathbf{d}^\top \mathbf{x} + s} \\\\
\textbf{s.t.} \quad  &\mathrm{C}_1:\quad \mathbf{x} \in \{0,1\}^n, \\
& \mathrm{C}_2: \quad \mathbf{A} \mathbf{x} \leq \mathbf{b}, 
\end{array}
\end{equation}
where $\mathbf{x} \in \{0,1\}^n$ is a binary decision vector, $\mathbf{Q}$ and $D$ are $n \times n$ symmetric matrices, $\mathbf{c}, \mathbf{d} \in \mathbb{R}^n$ are vectors, $r, s \in \mathbb{R}$ are constants, and $\mathbf{A} \in \mathbb{R}^{m \times n}, \mathbf{b} \in \mathbb{R}^m$ are optional linear constraint parameters.

 Also, $\left| \mathbf{a}^\top \mathbf{B} \right|^2$ can be written as

\begin{equation}
\begin{split}
    \left| \mathbf{a}^\top \mathbf{B} \right|^2 &= \mathbf{a}^\top \mathbf{B} \left(\mathbf{a}^\top \mathbf{B}\right)^H = \mathbf{a}^\top \mathbf{B} \mathbf{B}^H \mathbf{a} \\
    &= \mathbf{a}^\top \left(\Re\left(\mathbf{B} \mathbf{B}^H\right) + j\Im\left(\mathbf{B} \mathbf{B}^H\right)  \right) \mathbf{a} \\
    &= \mathbf{a}^\top \Re\left(\mathbf{B} \mathbf{B}^H\right) \mathbf{a} + j\mathbf{a}^\top \Im\left(\mathbf{B} \mathbf{B}^H\right) \mathbf{a}.
\end{split}
\end{equation}

Thus, since $\left| \mathbf{a}^\top \mathbf{B} \right|^2 \in \mathbb{R}$, it holds that 
\begin{equation}
\left| \mathbf{a}^\top \mathbf{B} \right|^2=\mathbf{a}^\top \Re\left(\mathbf{B} \mathbf{B}^H\right) \mathbf{a}. 
\end{equation}
Also, by setting $\mathbf{c} , \mathbf{r},\mathbf{D}, \mathbf{s}  = 0$, and $\mathbf{d} = 1$, it is concluded that \eqref{eq:max_snr} and \eqref{eq:our_qf01p} are equivalent.
\end{IEEEproof}

Interestingly, \( \mathbf{Q} \in \mathbb{R}^{N \times N} \) is a matrix that inherits several important properties from its construction:
\begin{itemize}
    \item \textbf{Real and symmetric:} By definition, \( \Re(\mathbf{B} \mathbf{B}^H) \) is a real-valued matrix. Furthermore, since \( \mathbf{B} \mathbf{B}^H \) is Hermitian, its real part is symmetric: \( \mathbf{Q}^\top = \mathbf{Q} \).
    
    \item \textbf{Positive semidefinite:} The Hermitian matrix \( \mathbf{B} \mathbf{B}^H \) is positive semidefinite, and the real part of a positive semidefinite Hermitian matrix remains positive semidefinite. Therefore, for any real vector \( \mathbf{a} \), it holds that \( \mathbf{a}^\top \mathbf{Q} \mathbf{a} \geq 0 \).
    
    \item \textbf{Low rank (rank at most 2):} Let \( \mathbf{r} = \Re(\mathbf{B}) \) and \( \mathbf{i} = \Im(\mathbf{B}) \) be the real and imaginary parts of \( \mathbf{B} \), respectively. Then,
    \begin{equation}
    \mathbf{Q} = \Re(\mathbf{B} \mathbf{B}^H) = \mathbf{r} \mathbf{r}^\top + \mathbf{i} \mathbf{i}^\top,
   \end{equation}
    which implies that \( \mathbf{Q} \) is a sum of two rank-one real symmetric matrices. Thus, \( \mathrm{rank}(\mathbf{Q}) \leq 2 \).
    
    \item \textbf{Interpretability:} Each element \( Q_{ij} \) measures the aligned signal strength between antennas \( i \) and \( j \), and is computed as
    \begin{equation}
    \mathbf{Q_{ij}} = \Re(\mathbf{B_i} \overline{\mathbf{B_j}}) = \Re(\mathbf{B_i}) \Re(\mathbf{B_j}) + \Im(\mathbf{B_i}) \Im(\mathbf{B_j}).
    \end{equation}
    This quantity reflects the real-valued correlation in signal contributions between the two antennas.
\end{itemize}

For additional properties of the 
\( \mathbf{Q} \) matrix, along with its relationship to the underlying complex vector 
\( \mathbf{B} \), refer to Appendix \ref{Appendix1}. The above observations clarify the role of the \( \mathbf{B} \)-values in our framework. The resulting problem \eqref{eq:our_qf01p}  is a special case of the QF01P problem, which is nonlinear, nonconvex, and NP-hard due to the binary constraint and the fractional structure of the objective. Notably, the matrix \( \mathbf{B} \) includes informative, physically grounded features that can be exploited by learning-based models to guide efficient and data-driven antenna selection strategies.

\section{A Deep Learning Based Solution}

Because this problem is NP-hard and realistic deployments are potentially high-dimensional, we explore deep learning-based approximations.
 To enable real-time antenna selection in scenarios where a brute-force search or other methods would be computationally prohibitive, we are exploring supervised learning models that approximate the optimal activation policy. Our goal is to learn a function that maps input features, such as user position and channel characteristics, to an activation vector, $\mathbf{a} \in \{0,1\}^N$, such that the selected antennas yield a near-optimal communication rate. We formulate this problem as a binary classification task where each antenna is either active (assigned a value of 1) or inactive (assigned a value of 0). The objective is to learn the activation probabilities for each antenna and select a configuration that yields an SNR close to optimal. These ``optimal" configurations are generated using the Gurobi solver, as detailed in the dataset generation section. Although Gurobi serves as a strong combinatorial optimization baseline, it is computationally expensive and unsuitable for real-time deployment, which further reinforces the practical value of our learning-based approach. In our pipeline, Gurobi is used exclusively for offline dataset creation, not for inference or decision-making.

We consider a fast heuristic method as a naive benchmark and propose three different deep learning architectures to solve the PAs optimization problem, which are described in detail below:
\begin{itemize}
    \item The nearest antenna method is a simple heuristic that selects for activation the antennas closest to the user in Euclidean distance. A fixed number of antennas are opened, prioritizing proximity without considering channel quality, phase alignment, or interference. This makes it fast and interpretable but fundamentally limited in multi-antenna environments.

    \item The first deep learning model is a simple MLP, designed to handle different number of antennas.
    \item The second model combines a GNN with an MLP, where the GNN first extracts relational features from the graph structure and the MLP processes the resulting node embeddings.
    \item The third model extends the second one by incorporating two attention layers, which allow the network to better capture complex interactions between antennas and improve its predictive capability. We refer to that as the GNN+distributed policy network (DisPN) model\footnote{The GNN+DisPN architecture was originally introduced to solve a variant of the Travelling Salesman Problem, another NP-hard combinatorial optimization problem \cite{DISPN}.}.
\end{itemize}  Through our experiments, we aim to use the heuristic method and the simpler architectures as baselines for more complex models. This allows us to demonstrate that stronger architectures are necessary to effectively address the task. We now proceed to analyze each architecture individually, highlighting their structural components and design motivations.

\subsection{\textbf{MLP}}
The first architecture is a MLP-based policy designed to process each user-antenna pair independently, while incorporating a form of global context aggregation. For each antenna \( n \) and user, the model receives as input the magnitude and phase of the complex channel coefficient \( B_n \), forming a 2-dimensional feature vector:
\begin{equation}
\mathbf{x}_n = [|B_n|, \angle B_n] \in \mathbb{R}^2.
\end{equation}
This input is passed through a shared local encoder (a one-layer MLP) to extract per-antenna features:
\begin{equation}
\mathbf{h}_n = \text{ReLU}(\mathbf{W}_1 \mathbf{x}_n + \mathbf{b}_1), \quad \mathbf{W}_1 \in \mathbb{R}^{d \times 2},\; \mathbf{h}_n \in \mathbb{R}^d.
\end{equation}

To introduce global context, the model computes the mean of all antenna features for a given user:
\begin{equation}
\bar{\mathbf{h}} = \frac{1}{N} \sum_{n=1}^{N} \mathbf{h}_n.
\end{equation}
Each local antenna feature is then concatenated with this global summary to form a fused representation:
\begin{equation}
\mathbf{z}_n = [\mathbf{h}_n \, \| \, \bar{\mathbf{h}}] \in \mathbb{R}^{2d}.
\end{equation}
The fused vector \( \mathbf{z}_n \) is passed through a deeper feedforward MLP, referred to as the fusion MLP, which produces the activation probability for antenna \( n \):
\begin{equation}
\hat{a}_n = \sigma(f_{\text{f}}(\mathbf{z}_n)),
\end{equation}
where \( \sigma(\cdot) \) is the sigmoid function and the function \( f_{\text{f}} \) is defined recursively as:

\begin{equation}
\begin{aligned}
&\mathbf{u}^{(0)} = \mathbf{z}_n, \\
&\mathbf{u}^{(l)} = \text{ReLU}(\mathbf{W}_l \mathbf{u}^{(l-1)} + \mathbf{b}_l), \quad \text{for } l = 1, \dots, L-1, \\
&f_{\text{f}}(\mathbf{z}_n) = \mathbf{W}_L \mathbf{u}^{(L-1)} + \mathbf{b}_L.
\end{aligned}
\label{eq:mlp_forward}
\end{equation}

Here, each \( \mathbf{W}_l \in \mathbb{R}^{d_{l} \times d_{l-1}} \) and \( \mathbf{b}_l \in \mathbb{R}^{d_l} \) are trainable parameters of the fusion MLP. In the implemented version, the fusion network consists of fully connected layers with ReLU activations, followed by a final sigmoid layer to produce scalar probabilities. While this architecture is simple and lightweight, it lacks explicit modeling of spatial or relational dependencies between antennas. As a result, it may underfit in complex scenarios where such interactions are critical for accurate antenna selection.

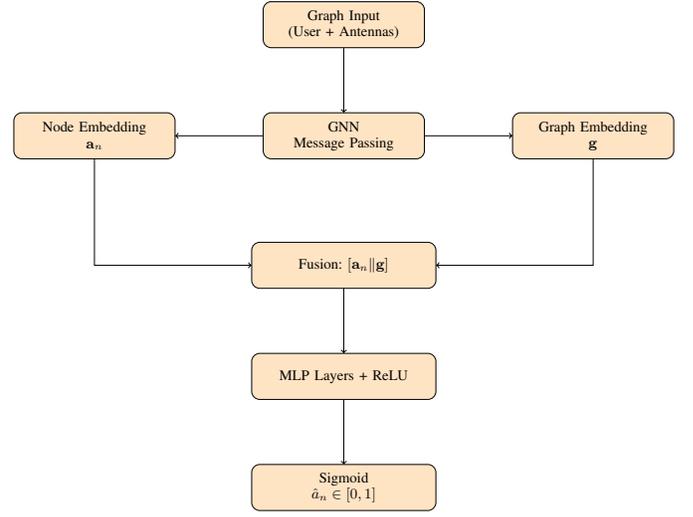
\begin{figure}[h]
\centering
\resizebox{\hsize}{!}{
\begin{tikzpicture}[
    node distance=1.4cm,
    every node/.style={font=\small, align=center},
    block/.style={draw, fill=peach, minimum width=3.5cm, minimum height=1cm, rounded corners=5pt},
    fusion/.style={draw, fill=peach, minimum width=4cm, minimum height=1cm, rounded corners=5pt},
    mlp/.style={draw, fill=peach, minimum width=4cm, minimum height=1cm, rounded corners=5pt},
    final/.style={draw, fill=peach, minimum width=4cm, minimum height=1cm, rounded corners=5pt}
]

\node[block] (input) {Graph Input \\ (User + Antennas)};
\node[block, below=of input] (gnn) {GNN \\ Message Passing};
\node[block, left=1.9cm of gnn] (nodeemb) {Node Embedding \\ $\mathbf{a}_n$};
\node[block, right=1.9cm of gnn] (graphemb) {Graph Embedding \\ $\mathbf{g}$};
\node[fusion, below=1.8cm of gnn] (fusion) {Fusion: $[\mathbf{a}_n \| \mathbf{g}]$};
\node[mlp, below=of fusion] (mlp) {MLP Layers + ReLU};
\node[final, below=of mlp] (output) {Sigmoid \\ $\hat{a}_n \in [0, 1]$};

\draw[->] (input) -- (gnn);
\draw[->] (gnn) -- (nodeemb);
\draw[->] (gnn) -- (graphemb);
\draw[->] (nodeemb.south) |- (fusion.west);
\draw[->] (graphemb.south) |- (fusion.east);
\draw[->] (fusion) -- (mlp);
\draw[->] (mlp) -- (output);

\end{tikzpicture}}
\caption{Architecture overview of the GNN + MLP model for antenna activation prediction.}
\label{fig:gnn_mlp_architecture}
\end{figure}

\subsection{\textbf{GNN + MLP}}

\begin{figure}[h]
\resizebox{.95\hsize}{!}{
\begin{tikzpicture}
\tikzstyle{mainnode} = [circle, draw, fill=peach, minimum size=1.2cm, font=\small]
\node[mainnode] (user) at (0, -3.5) {$\psi_m$};
\node[mainnode] (a1) at (-5, 0) {$\psi^{\text{Pin}}_0$};
\node[mainnode] (a2) at (-2.5, 0) {$\psi^{\text{Pin}}_1$};
\node[mainnode] (a3) at (0, 0) {$\psi^{\text{Pin}}_2$};
\node[mainnode] (a4) at (2.5, 0) {$\psi^{\text{Pin}}_3$};
\node[mainnode] (an) at (5, 0) {$\psi^{\text{Pin}}_n$};
\draw[-] (a1) -- (user);
\draw[-] (a2) -- (user);
\draw[-] (a3) -- (user);
\draw[-] (a4) -- (user);
\draw[-] (an) -- (user);
\node at (-4.5, -1) {\scriptsize $[|B_0|, \angle B_0]$};
\node at (-2.6, -1) {\scriptsize $[|B_1|, \angle B_1]$};
\node at (-0.85, -1) {\scriptsize $[|B_2|, \angle B_2]$};
\node at (1, -1) {\scriptsize $[|B_3|, \angle B_3]$};
\node at (4.8, -1) {\scriptsize $[|B_{N-1}|, \angle B_{N-1}]$};
\node at (3.5, 0) {\scriptsize $\cdot$};
\node at (3.7, 0) {\scriptsize $\cdot$};
\node at (3.9, 0) {\scriptsize $\cdot$};
\end{tikzpicture}
}
\caption{Graph representation of a user connected to $n$ PAs.}
\label{fig:pinching_graph}
\end{figure}
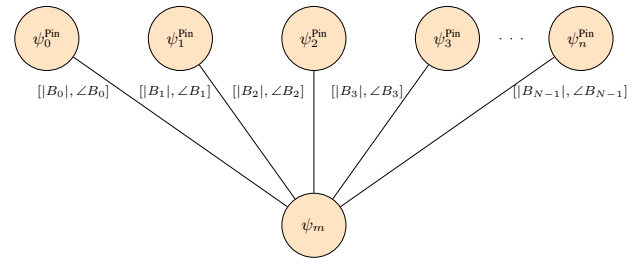

The second architecture, which is presented in Fig. \eqref{fig:gnn_mlp_architecture}, augments the MLP-based policy by incorporating a GNN to explicitly model spatial and relational dependencies between antennas and the user. As it is shown in Fig. \ref{fig:pinching_graph}, this is achieved by formulating each instance as a graph where the nodes represent the antennas and the user, and the edges encode the complex-valued channel coefficients between them.

\paragraph*{Graph Construction}
For a given instance with \( N \) antennas and one user, we construct a  graph with \( N + 1 \) nodes: one node for each antenna and one node (the first one) for the user. An edge is added between the user and each antenna. The edge features correspond to the complex channel coefficient \( B_n \) between user and antenna \( n \), expressed in polar form as:
\[
\text{Edge feature: } \mathbf{e}_{n} = [|B_n|, \angle B_n] \in \mathbb{R}^2.
\]

\subsubsection{Node and Graph Embeddings}
The GNN backbone consists of a stack of Message Passing Neural Network (MPNN) layers, where node representations are iteratively updated through edge-conditioned message aggregation. Initial node features are projected to the hidden dimension via a linear transformation, followed by message-passing layers that incorporate edge-aware updates. At each layer \( l \), the node embedding \( \mathbf{h}_v^{(l)} \) for node \( v \) is updated as:
\begin{equation}
\mathbf{h}_v^{(l)} = \text{ReLU} \left( \sum_{u \in \mathcal{N}(v)} \text{f}^{(l)}(\mathbf{e}_{uv}) + \mathbf{b}_v^{(l)} \right),
\end{equation}
where \( \mathbf{e}_{uv} \) denotes the edge feature between nodes \( u \) and \( v \), and \( \mathcal{N}(v) \) is the neighborhood of node \( v \). The final node embeddings are obtained by aggregating representations across all layers. In particular, the embeddings corresponding to the \( N \) antenna nodes are extracted and used as input to downstream modules.

To obtain a graph-level representation, we perform both mean and max pooling over the node embeddings at each layer, followed by layer-specific transformations. The resulting pooled vectors are summed to produce the final graph embedding. Specifically, for each layer \( l \in \{0, \dots, L\} \), we concatenate the mean and max pooled node features from \( \mathbf{H}^{(l)} \), the matrix of node embeddings at layer \( l \), and pass them through a learnable transformation \( \text{f}_{(l)} \). Here, \( \mathbf{H}^{(l)} \in \mathbb{R}^{N \times d} \) denotes the matrix of node embeddings at layer \( l \), where each row corresponds to the embedding \( \mathbf{h}_v^{(l)} \) of node \( v \):
\begin{equation}
\mathbf{H}^{(l)} =
\begin{bmatrix}
\mathbf{h}_1^{(l)} \\
\mathbf{h}_2^{(l)} \\
\vdots \\
\mathbf{h}_N^{(l)}
\end{bmatrix}.
\end{equation}
The final graph embedding is then obtained by summing the transformed representations from all layers:
\begin{equation}
\mathbf{g} = \sum_{l=0}^{L} \text{f}_{(l)}\left( \left[ \text{mean-pool}(\mathbf{H}^{(l)}), \, \text{max-pool}(\mathbf{H}^{(l)}) \right] \right).
\end{equation}

\subsubsection{Fusion and Prediction}
Let \( \mathbf{a}_n \in \mathbb{R}^{d} \) denote the final embedding of antenna \( n \), and \( \mathbf{g} \in \mathbb{R}^{d} \) the global context vector. For each antenna, we concatenate its local embedding with the global context:
\begin{equation}
\mathbf{z}_n = [\mathbf{a}_n \, \| \, \mathbf{g}] \in \mathbb{R}^{2d},
\end{equation}
and pass it through a feedforward MLP (eq. \ref{eq:mlp_forward}) to produce activation logits:
\begin{equation}
\hat{a}_n = \sigma(f(\mathbf{z}_n)),
\end{equation}

Here, the sigmoid activation maps the output into the interval \([0,1]\), and the weights \( \{\mathbf{W}_l, \mathbf{b}_l\} \) define the fusion MLP. This architecture improves over the plain MLP by capturing the geometric and pairwise interactions among antennas and the user, allowing the model to learn more expressive features relevant to optimal antenna selection.

\subsection{\textbf{GNN + Distributed Policy Network (DisPN)}}

\begin{figure}[h]
    \centering
    \includegraphics[width=1\linewidth]{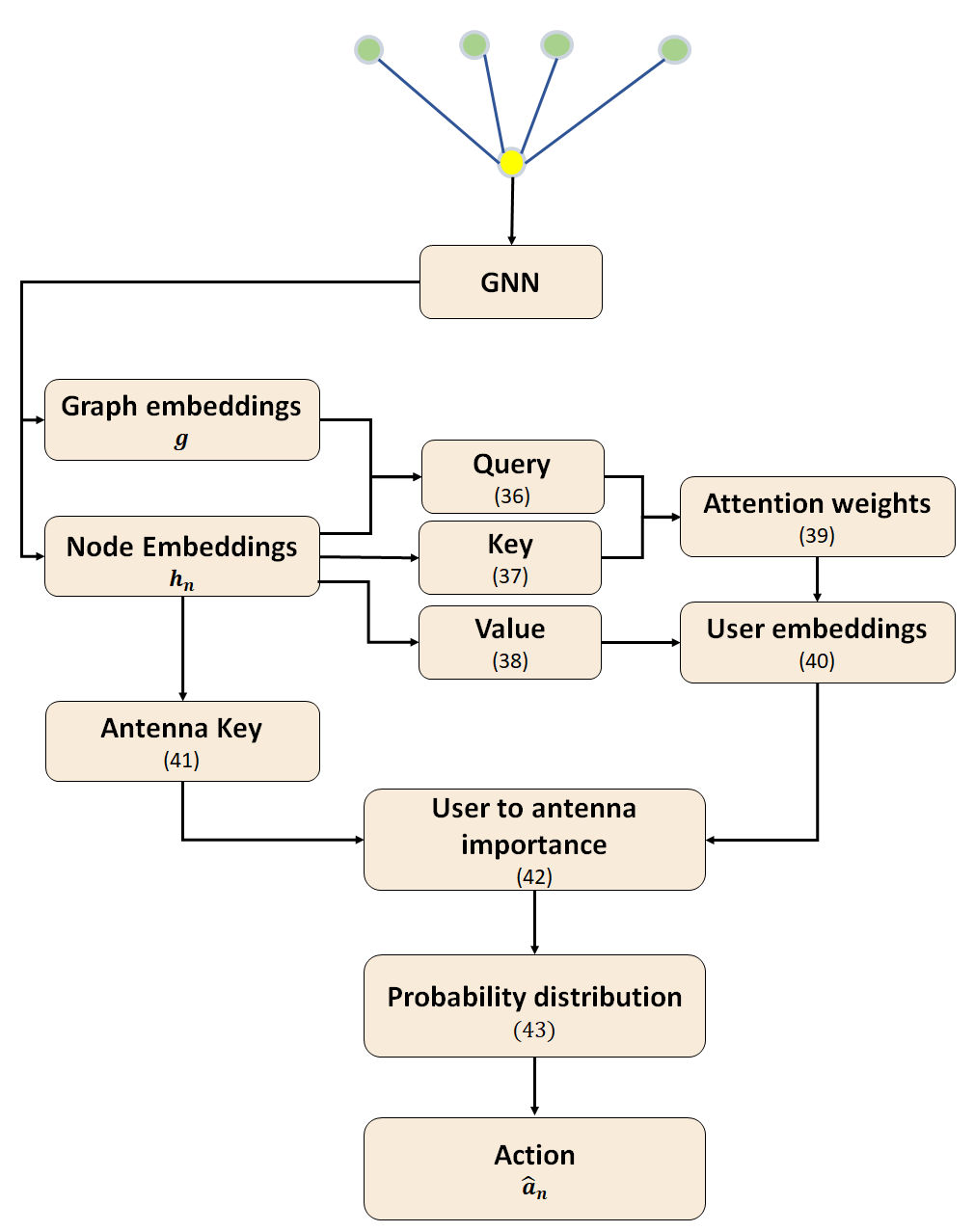}
    \caption{GNN-DisPN architecture}
    \label{fig:enter-label}
\end{figure}

The most advanced architecture we propose enhances the GNN-MLP design with a distributed attention mechanism that enables the model to reason dynamically about antenna importance in a user-aware and context-dependent manner, as it is shown in Fig. \ref{fig:enter-label}. The pipeline involves three key stages: (i) relational feature extraction via a message-passing GNN, (ii) construction of a user-specific embedding through cross-attention, and (iii) final antenna activation via attention-based scoring.

\subsubsection{Distributed Attention Mechanism}

To generate contextual decisions, the user embedding is treated as a query source, while each antenna embedding acts as a key. The attention mechanism operates in two main steps:

    \paragraph{Agent Embedding}  
    To construct a context-aware representation for the user (agent), we concatenate the global graph embedding $\mathbf{g} \in \mathbb{R}^d$ with the local embedding of the user node (typically the first node in the graph). This concatenated vector is projected into a query vector:
   \begin{equation}
        \mathbf{q} = \mathbf{W}_q [\mathbf{g}; \mathbf{h}_{\text{user}}] \in \mathbb{R}^{d_k},
    \end{equation}
    where \( d_k \in \mathbb{N} \) denotes the dimensionality of the key and query vectors used in the attention mechanism and $\mathbf{W}_q \in \mathbb{R}^{(2d) \times d_k}$ . Each antenna embedding $\mathbf{h}_n \in \mathbb{R}^d$ is projected into key and value representations:
    \begin{equation}
        \mathbf{k}_n = \mathbf{W}_k \mathbf{h}_n, 
    \end{equation}
      \begin{equation}
     \mathbf{v}_n = \mathbf{W}_v \mathbf{h}_n,
    \end{equation}
    with $\mathbf{W}_k, \mathbf{W}_v \in \mathbb{R}^{d \times d_k}$. The attention weight for each antenna is computed as:
    \begin{equation}
        w_n = \sigma\left( \frac{\mathbf{q}^\top \mathbf{k}_n}{\sqrt{d_k}} \right),
    \end{equation}
    and the resulting user embedding is the weighted sum of the value vectors:
    \begin{equation}
        \mathbf{z}_{\text{u}} = \sum_n w_n \mathbf{v}_n.
    \end{equation}

    \paragraph{Policy Attention and Importance Scores}  
    The agent embedding $\mathbf{z}_{\text{u}}$ is then used to compute importance scores for each antenna. First, each antenna embedding is projected again:
    \begin{equation}
        \mathbf{k}'_n = \mathbf{W}_k' \mathbf{h}_n,
    \end{equation}
    where $\mathbf{W}_k' \in \mathbb{R}^{d \times d_k}$. The importance score \( \text{imp}_n \in \mathbb{R} \) for each antenna is given by:
    \begin{equation}
        \text{imp}_n = c \cdot \tanh\left( \frac{\mathbf{z}_{\text{u}}^\top \mathbf{k}'_n}{\sqrt{d_k}} \right),
    \end{equation}
    where \( c > 0 \) is a sharpening coefficient. Finally, the activation probability for antenna \( n \), which determines \(\hat{a}_n \), is obtained by applying a sigmoid to the importance:
    \begin{equation}
        \hat{p}_n = \sigma(\text{imp}_n).
    \end{equation}
    This two-step formulation—computing importance scores followed by a sigmoid—provides precise control over activation sharpness and enables interpretation of each antenna's relative importance to the user. 

\subsubsection{Learned Parameters}
The following weight matrices are learned during training:
\begin{itemize}
    \item $\mathbf{W}_q$: projects the concatenated user and global graph features into query space.
    \item $\mathbf{W}_k$, $\mathbf{W}_v$: project antenna node features into key/value spaces for agent attention.
    \item $\mathbf{W}_k'$: projects antenna features for final policy scoring.
\end{itemize}

\subsubsection{Advantages and Capabilities}

This attention-based formulation provides several advantages: (i) it enables distributed reasoning where the user embedding dynamically interacts with all antennas; (ii) it allows the model to focus selectively on antennas based on both spatial configuration and signal properties; (iii) it produces sharper activation decisions via the sharpening parameter $c$. In addition, the model naturally supports interpretability and explainability, as the attention scores explicitly reflect the relative importance of each antenna to the user. This is a valuable property of the architecture, as interpretability in real-time communication systems is both challenging and crucial. As a result, the GNN+DisPN model is particularly well-suited for large-scale or noisy deployments where fine-grained coordination is necessary.

\subsection{Dataset Generation}
To train the deep learning models, we generate a dataset consisting of optimal solutions for the PAs problem. We focus on a configuration with 50 antennas, where for each instance, the corresponding matrix $\mathbf{Q} \in \mathbb{R}^{50 \times 50}$ is constructed based on the physical parameters of the system. To obtain the ground-truth optimal solutions, we use Gurobi optimizer and apply Dinkelbach's algorithm, a well-known iterative method for solving fractional programming problems \cite{dinkelbach1967nonlinear}. For more details about this method, refer to Appendix \ref{Appendix2}.  Each dataset sample includes the user position and the corresponding complex channel coefficients \( \mathbf{B} \),  which are necessary in order to construct the graph representation of the problem  and find the optimal activation vector \( \mathbf{a}^\star \). This dataset serves as the supervised learning basis for training the aforementioned deep learning models providing the instances with their labeled solution.
We use a training dataset with 5000 instances and a validation dataset of 1000 instances which is generated using the same procedure. It is used during training to monitor the model's performance and prevent overfitting. To evaluate the generalization ability of the model, we also prepare a test dataset consisting of problem instances with a larger number of antennas, specifically 100, 200, 500 and 1000 antennas. All datasets are generated under the same spatial configuration assumptions described in Table \ref{tab:system_params}.

\subsection{Custom Loss Function}

At first, the models are trained using the binary cross-entropy (BCE) loss between the predicted activation probabilities \( \hat{a}_n \in [0,1] \) and the ground truth binary activations \( a_n^\star \in \{0,1\} \). For a single instance (i.e., one user-antenna configuration), the loss is defined as:
\begin{equation}
\mathcal{L}_{\text{BCE}} = - \sum_{n=1}^N \left[ a_n^\star \log(\hat{a}_n) + (1 - a_n^\star) \log(1 - \hat{a}_n) \right],
\end{equation}
where \( N \) is the number of antennas.  This formulation encourages each model to independently predict accurate binary activation decisions for all antennas, guided by optimal data generated via combinatorial solvers. 

It is important to note that the training data exhibits a strong class imbalance, as the majority of antennas are typically inactive (i.e., closed) in the optimal solutions provided by the combinatorial solvers. This bias toward the negative class leads to a skewed label distribution, where the model may achieve low BCE loss simply by predicting most antennas as closed. However, such behavior can obscure the model’s ability to correctly identify the few antennas that should be active. As BCE loss treats each activation decision independently and equally, it may not sufficiently penalize false negatives in this context. This makes standard BCE potentially suboptimal for learning in the presence of this imbalance, motivating the need to explore alternative loss functions or class reweighting schemes that better reflect the asymmetry in the label distribution and the importance of correctly predicting active antennas.

We can use an augmented custom loss function $\mathcal{L}_c$ that combines multiple objectives. The foundation of this formulation is a weighted BCE (WBCE) loss, computed on the model's raw importance scores, denoted by $\text{imp}$. To address the class imbalance, the positive class (i.e., active antennas) is upweighted using a scalar factor \( \alpha > 1 \), which also helps prevent collapse into trivial all-zero solutions.

Formally, this loss is given by:

\begin{equation}
\scalebox{0.85}{$
\mathcal{L}_{\text{WBCE}} = - \frac{1}{N} \sum_{i=1}^{N} \left[ \alpha \cdot y_i \cdot \log \sigma(\text{imp}_i) + (1 - y_i) \cdot \log (1 - \sigma(\text{imp}_i)) \right],
$}
\end{equation}
where \( y_i \in \{0, 1\} \) is the ground truth activation for the $i$-th antenna.

On top of this classification loss, we incorporate signal-aware terms that promote end-to-end performance. Specifically, soft antenna activations are obtained by applying a sigmoid function to the model’s output logits, and are then used to estimate the resulting SNR \( \hat{\gamma} \) via a differentiable approximation of the system-level rate computation. We introduce a penalty term that minimizes the squared deviation from the optimal SNR \( \gamma^\star \):

\begin{equation}
\mathcal{L}_{\gamma} = \left(1 - \frac{\hat{\gamma}}{\gamma^\star}\right)^2,
\end{equation}
which guides the model toward high-performing configurations beyond binary correctness alone.

Additionally, to prevent collapse into low-SNR regimes, we include a hinge-style regularization term that penalizes activations yielding significantly suboptimal SNR:
\begin{equation}
\mathcal{L}_{\text{C}} = \text{ReLU}\left(0.1 - \frac{\hat{\gamma}}{\gamma^\star}\right).
\end{equation}

The overall loss function is given by a weighted sum of these components:
\begin{equation}
    \mathcal{L} = \lambda_1 \cdot \mathcal{L}_{\text{WBCE}} + \lambda_\gamma \cdot \mathcal{L}_{\text{SNR}} + \lambda_c \cdot \mathcal{L}_{\text{C}},
    \label{eq:loss_function}
\end{equation}
where $\lambda_1$, $\lambda_\gamma$, and $\lambda_c$ are tuning parameters.
During training, the total loss is computed as the mean BCE loss over all instances in the batch:
\begin{equation}
\mathcal{L}{\text{oss}} = \frac{1}{B} \sum_{b=1}^B \mathcal{L}^{(b)},
\end{equation}
where \( B \) is the batch size and \( \mathcal{L}^{(b)} \) denotes the loss for the \(b\)-th instance. This composite loss allows the model to balance binary correctness with end-to-end signal quality, leading to more robust and effective antenna selection policies.

\section{Extension and Evaluation under User Location Uncertainty}

In practical deployment scenarios, the exact user position is typically unknown due to localization errors. To address this, we adopt a Monte Carlo inference approach that simulates real-time uncertainty by generating multiple noisy estimates of the user’s location.

Let $\boldsymbol{\psi}$ denote the true user position. At inference time, we sample $M$  estimates of the user position:
\begin{equation}
\boldsymbol{\psi}^{(i)} = \boldsymbol{\psi} + \boldsymbol{\epsilon}^{(i)}, \quad \boldsymbol{\epsilon}^{(i)} \sim \mathcal{N}(\boldsymbol{0}, \sigma_p^2 \mathbf{I}_3), \quad i = 1, \dots, M,
\end{equation}
where $\sigma_p^2$ is the variance of the position estimation errors.

For each sample $\boldsymbol{\psi}^{(i)}$, we compute the corresponding complex channel vector $\mathbf{B}^{(i)}$ and construct a graph representation, which is passed through the trained policy network to obtain the predicted activation probabilities:
\begin{equation}
\boldsymbol{\pi}^{(i)} = \text{PolicyNet}(\mathbf{G}^{(i)}),
\end{equation}
where $\mathbf{G}^{(i)}$ is the graph constructed from the noisy channel features. To derive a robust final activation decision, we aggregate the probability predictions across all $M$ samples using the following method, hereinafter termed as mean activation policy.
\paragraph*{Mean Activation Policy}
To produce a robust binary activation decision for each antenna, we aggregate the activation probabilities across a set of sampled policy outputs. Specifically, we compute the element-wise mean of the activation vectors:
\begin{equation}
\bar{\boldsymbol{\pi}} = \frac{1}{M} \sum_{i=1}^M \boldsymbol{\pi}^{(i)},
\end{equation}
where \( \boldsymbol{\pi}^{(i)} \in [0,1]^N \) denotes the probability vector produced by the policy in the \( i \)-th forward pass, and \( M \) is the number of such passes (e.g., due to sampling in a stochastic policy). Each entry \( \bar{\pi}_j \) in the averaged vector \( \bar{\boldsymbol{\pi}} \) represents the mean probability of antenna \( j \) being activated across samples.

To obtain a final binary decision vector \( \hat{\mathbf{a}} \in \{0,1\}^N \), we apply a thresholding rule:
\begin{equation}
\hat{\mathbf{a}} = \mathbb{I}(\bar{\boldsymbol{\pi}} > 0.5),
\end{equation}
where \( \mathbb{I}(\cdot) \) denotes the element-wise indicator function. Intuitively, this means that an antenna is selected (i.e., activated) in the final output if it is more frequently predicted as active than inactive across the sampled policies. In other words, antennas that are more often considered useful (i.e., "open") than not are ultimately turned on in the final configuration. This procedure ensures that antenna activations are stable under realistic levels of user location noise and improves robustness to localization errors during deployment.

\section{Simulation Results}
In this section, we present the performance results for a nearest antenna heuristic, the MLP with fusion layer, the GNN-MLP model, and the GNN+DisPN model. We report classification accuracies for each architecture on the test sets, as well as additional comparison metrics. These results provide a comprehensive evaluation of each method's ability to predict optimal antenna activation patterns. 

\begin{table}[h!]
\centering
\caption{System Model Parameters.}
\label{tab:system_params}
\begin{tabular}{@{}ll@{}}
\toprule
\textbf{Parameter} & \textbf{Value} \\
\midrule
Number of users \(n \)  & 1 \\
Number of pinches \( N \) &   50--1000 \\
Carrier frequency \( f_c \) & 3~GHz  \\
Transmit SNR $ \rho = \frac{\eta P}{\sigma^2_{w}}$ &  40dB \\
Pinch height \( H \) & 3m  \\
Square area side length $2L$ & 10m \\
Waveguide length  $2D$  & 5m  \\
\bottomrule
\end{tabular}
\end{table}

All models were trained with BCE loss for 5000 iterations with hyperparameters shown in Table \ref{tab:system_params}. They were tested on configurations with 50 antennas, and then additional experiments assessed their ability to generalize to larger-scale deployments. 
The performance comparison for the 50-antenna case is summarized in Table~\ref{tab:50ant_snr}. The baseline MLP model achieves 82\% test accuracy in terms of received SNR and 82\% 
accuracy in terms of achievable rate. Incorporating structural information via the GNN+MLP model results in a clear improvement, raising SNR accuracy to 86\%. Our proposed architecture, GNN+DisPN, achieves the best overall performance, reaching 87\% SNR accuracy and 85\%  accuracy in terms of achievable rate on the test set.
While GNN+DisPN is more computationally involved, its complexity remains within practical bounds. As shown in Table~\ref{tab:complexity}, its forward pass time in our computer using CPU) is only slightly higher (1.7 ms vs. 1.5 ms for GNN+MLP and 0.5 ms for the MLP), and the number of parameters increases to 120 k compared to 50 k for the other two models. However, the increased  complexity is justified by a substantial gain in generalization performance, as demonstrated in the following evaluations. Moreover, despite the higher parameter count, the increase in floating point operations (FLOPs ) is modest (9.78 M vs. 5.48 M), indicating that GNN+DisPN remains efficient and scalable for real-time inference.

\begin{table}[h!]
\centering
\caption{Comparison of Model Performances for 50 antennas.}
\label{tab:50ant_snr}
\resizebox{\hsize}{!}{
\begin{tabular}{@{}lcccc@{}}
\toprule
Model & Train Acc (SNR) & Test Acc (SNR) & Test Acc (Bitwise)\\
\midrule
MLP & 82 \% & 82 \% & 82 \% \\
GNN + MLP & 86 \% & 86 \% & 82 \%\\
GNN + DisPN & 87 \% & 87 \% & 85 \%  \\
\bottomrule
\end{tabular}}
\end{table}

\begin{table}[h!]
\centering
\caption{Comparison of Model Complexities for 50 antennas}
\label{tab:complexity}
\resizebox{\linewidth}{!}{%
\begin{tabular}{@{}lcccc@{}}
\toprule
Model & Forward pass time (ms) & Number of parameters (k) & FLOPS (M) \\
\midrule
MLP & 0.5 & 50 & 5.06 \\
GNN + MLP & 1.5 & 50 & 5.48 \\
GNN + DisPN & 1.7 & 120 & 9.78 \\
\bottomrule
\end{tabular}%
}
\end{table}

To assess generalization, we evaluate the models—trained solely on 50-antenna scenarios—on test configurations with 100, 200, 500, and 1000 antennas. As shown in Table~\ref{tab:more_antennas_snr}, the MLP fails to generalize, yielding SNR accuracies near 65\% and inconsistent rate predictions. GNN+MLP maintains reasonable performance across scales, but it is the GNN+DisPN model that demonstrates exceptional scalability, consistently achieving over 90\% SNR accuracy. This confirms the model's capacity to generalize beyond the training distribution and operate effectively in larger antenna systems. For comparison, we also include the nearest-antenna method, which activates as many antennas as activated in the GNN + DisPN model, but the ones closest to the user. This naive baseline achieves merely 7\% test accuracy, as it ignores critical factors such as constructive interference and spatial diversity. By selecting antennas based purely on distance, it fails to account for phase alignment and channel gain variations, leading to severely degraded signal quality.

\begin{table}[h!]
\centering
\caption{Comparison of Model Performances in bigger number of antennas (SNR accuracies).}
\label{tab:more_antennas_snr}
\begin{tabular}{@{}lcccc@{}}
\toprule
Model & 100 & 200 & 500 & 1000\\
\midrule
Nearest Antennas & 7 \% & 6 \% & 6 \% & 7 \%\\
MLP & 66 \% & 64 \% & 68 \% & 65 \% \\
GNN + MLP & 81 \% & 80 \% & 81 \% & 79 \%\\
GNN + DisPN & 91 \% & 94 \% & 95 \% & 94 \%\\
\bottomrule
\end{tabular}
\end{table}

\begin{figure}[!t]
\centering
\begin{tikzpicture}
    \begin{axis}[
        width=1\linewidth,
        xlabel={Iterations},
        ylabel={Accuracy},
        ymin=0,
        ymax=1,
        xmin=0,
        xmax=5000,
        grid=major,
        legend entries = {
            {Validation Accuracy},
            {Training Accuracy},
        },
        legend style={font=\small, at={(0.2,0)}, anchor=south west},
    ]
    \addplot[blue,  line width=1pt]
    table {Val_acc_converge.dat};
    \addplot[red, line width=1pt]
    table {Train_acc_converge.dat};
    \end{axis}
\end{tikzpicture}
\vspace{-0.1in}
\caption{Accuracy Convergence.}
\label{fig:Convergence_plot}
\vspace{-0.15in}
\end{figure}
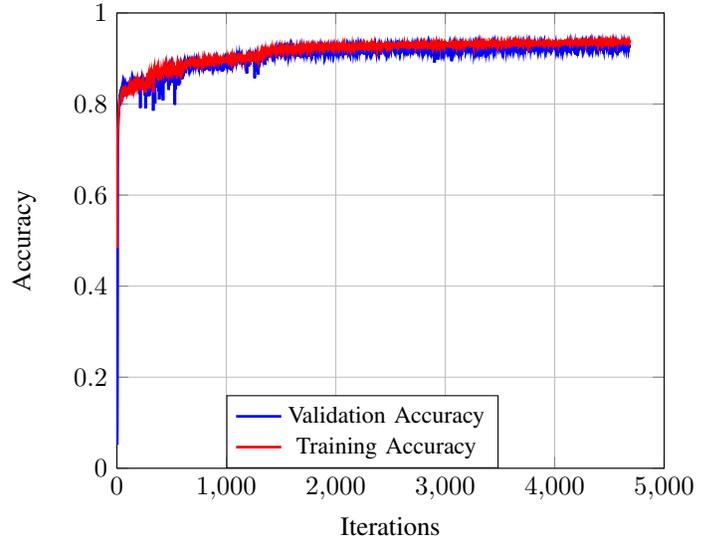

Given that the GNN+DisPN architecture achieved the highest accuracy among the evaluated models, we further focused our investigation on this model. To improve its robustness and performance, we trained it with the augmented loss function presented earlier in Section III-\eqref{eq:loss_function}, which combines a weighted binary cross-entropy term with signal-aware penalties based on soft SNR estimation and collapse avoidance. Using this augmented loss, the GNN+DisPN model achieves a notable improvement, increasing its SNR accuracy on the 50-antenna test set from 87\% to 93\%. The convergence curve of this model is shown in Fig. \ref{fig:Convergence_plot}. No overfitting was observed, as the model efficiently reaches similar training and validation accuracy. The hyperparameters used for this training are shown in Table \ref{tab:hyperparams}.

\begin{table}[h!]
\centering
\caption{Training Hyperparameters.}
\label{tab:hyperparams}
\begin{tabular}{@{}lc@{}}
\toprule
\textbf{Hyperparameter} & \textbf{Value} \\
\midrule
Batch size & 1000 \\
Learning rate & $10^{-4} \xrightarrow{} 10^{-5}$ \\
User \& Node embedding size & 128 \\
Attention's key and value size & 64 \\
Number of GNN layers & 1 \\
Clipping constant c & 10 \\
Loss coefficient \( \alpha \) &  1.6\\
Loss coefficient \( \lambda_1 \) &  $0.5 \xrightarrow{} 0.3$\\
Loss coefficient \( \lambda_\gamma \) & $2 \xrightarrow{} 8$ \\
Loss coefficient \( \lambda_c \) & $100 \xrightarrow{} 20$ \\
\bottomrule
\end{tabular}
\end{table}

\subsection{K Selection Method}
We explore scenarios in which only a subset of the predicted antennas is activated through a post-processing strategy. Specifically, given a model prediction that selects \( N \) antennas, we apply a \( K \)-antenna refinement step, where the \( K \) antennas with the highest predicted importance scores are retained. The value of \( K \) is defined as a fixed fraction of \( N \), i.e., \( K = kN \), with \( k \in (0,1] \).

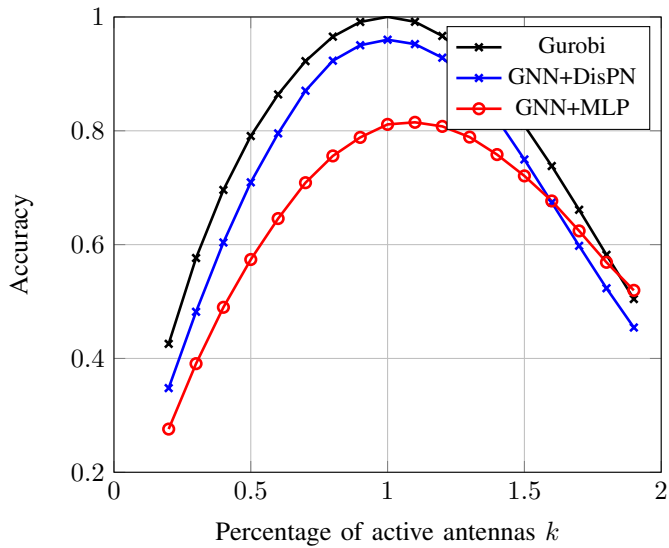
\begin{figure}[!t]
\centering
\begin{tikzpicture}
    \begin{axis}[
        width=1\linewidth,
        xlabel={Percentage of active antennas $k$},
        ylabel={Accuracy},
        ymin=0.2,
        ymax=1,
        xmin=0,
        xmax=2,
        grid=major,
        legend entries = {
            {Gurobi},
            {GNN+DisPN},
            {GNN+MLP}
        },
        legend style={font=\small, at={(0.98,0.98)}, anchor=north east},
    ]
    \addplot[black, mark=x, line width=1pt]
    table {snr_acc_Gurobi_KSelection.dat};
    \addplot[blue, mark=x, line width=1pt]
    table {snr_acc_GNN_DispN_KSelection.dat};
    \addplot[red, mark=o, line width=1pt]
    table {snr_acc_GNN_MLP_KSelection.dat};
    \end{axis}
\end{tikzpicture}
\vspace{-0.1in}
\caption{Top-$K$ selection method for 500 antennas.}
\label{fig:TOPK_500_selection}
\vspace{-0.15in}
\end{figure}

Fig. ~\ref{fig:TOPK_500_selection} illustrates the effect of the top-$K$ selection strategy on the mean SNR ratio for a system with 500 antennas. As expected, the SNR ratio peaks when the number of activated antennas matches the model’s original selection ( i.e., $k= 1.0$ ), representing the best-case performance according to each model’s internal ranking.
When fewer antennas are selected ($k < 1$), performance declines due to insufficient activation, since important antennas are excluded, resulting in reduced received power. Conversely, when more antennas than necessary are activated (\(k > 1\)), the performance also deteriorates due to \textit{phase misalignment}: the additional antennas may introduce signals that are not coherently aligned, leading to destructive interference that lowers the overall SNR. Moreover, since the SNR expression includes the number of active antennas in the denominator, activating antennas that contribute weakly or incoherently can further degrade performance by reducing the effective signal strength and amplifying the impact of noise.

Across all values of $k$, the GNN+DisPN model maintains a higher SNR ratio compared to GNN+MLP, demonstrating its superior ability to rank antennas and maintain robustness under varying selection thresholds. The GNN+MLP model, lacking an explicit attention mechanism, struggles to assign accurate importance scores to individual antennas, resulting in suboptimal selections.
In comparison to the Gurobi-based optimal solution, GNN+DisPN closely approaches optimal performance near the model's original selection point. While a performance gap remains, GNN+DisPN effectively bridges the space between learned heuristics and exact combinatorial optimization, unlike GNN+MLP, which exhibits a consistent lag in SNR ratio across all selection percentages. Notably, for selection ratios \( k \in [0.7, 1.0] \), the SNR accuracy of GNN+DisPN remains above 90\% of the optimal. This demonstrates that the model can use some of its strength while still maintaining near-optimal signal quality.

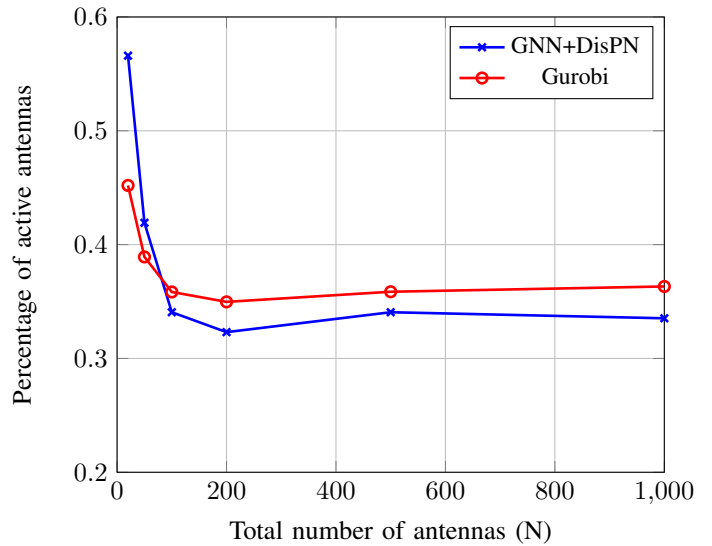
\begin{figure}[!t]
\centering
\begin{tikzpicture}
    \begin{axis}[
        width=1\linewidth,
        xlabel={Total number of antennas (N)},
        ylabel={Percentage of active antennas},
        ymin=0.2,
        ymax=0.6,
        xmin=0,
        xmax=1000,
        grid=major,
        legend entries = {
            {GNN+DisPN},
            {Gurobi},
        },
        legend style={font=\small, at={(0.98,0.98)}, anchor=north east},
    ]
    \addplot[blue, mark=x, line width=1pt]
    table {n_ant_test_vs_avg_activation_model.dat};
    \addplot[red, mark=o, line width=1pt]
    table {n_ant_test_vs_avg_activation_opt.dat};
    \end{axis}
\end{tikzpicture}
\vspace{-0.1in}
\caption{Antennas activation percentage.}
\label{fig:ActivationRatio}
\vspace{-0.15in}
\end{figure}

Another important aspect to analyze is how the fraction of activated antennas evolves as the system scales. Fig.~\ref{fig:ActivationRatio} illustrates the percentage of active antennas, which is computed as the number of selected antennas divided by the total number \( N \), for both the model and the optimal Gurobi-based solution, across increasing antenna counts. As \( N \) grows, both curves exhibit a decreasing trend that stabilizes beyond 200 antennas and converges to a percentage of 0.37. This reflects a natural system behavior: activating more antennas spreads the available transmit power more thinly and increases the risk of phase misalignment. As a result, both the model and the optimal policy adopt a more selective activation strategy as the antenna array grows, converging to a stable activation ratio that balances power efficiency and signal coherence.

\subsection{Localization Uncertainty}
We now investigate the model's ability to handle uncertainty in user localization, which is a critical challenge in practical deployments. The resulting activation probabilities are aggregated across positions estimates samples around the real user position to produce a final robust activation decision. As described earlier, the aggregation can be performed using the mean activation policy. In this setting, the uncertainty level is governed by the standard deviation parameter \( \sigma_p \), which directly controls the noise in the user's position. For example, a value of \( \sigma_p = 0.3 \) corresponds to an average positional error of approximately \( 2\lambda \), where \( \lambda \) is the free-space wavelength. Such spatial deviations introduce significant phase shifts and channel mismatches, resulting in a realistic and challenging communication environment. 

\begin{figure}[!t]
\centering
\begin{tikzpicture}
    \begin{axis}[
        width=1\linewidth,
        xlabel={$\sigma_p$},
        ylabel={Accuracy},
        ymin=0.5,
        ymax=1,
        xmin=0,
        xmax=0.4,
        grid=major,
        legend entries = {
            {GNN+DisPN },
            {Gurobi},
        },
        legend style={font=\small, at={(0.2,0)}, anchor=south west},
    ]
    \addplot[blue,  line width=1pt]
    table {snr_acc_model_noise.dat};
    \addplot[red, line width=1pt]
    table {snr-acc_gurobi_noise.dat};
    
    \end{axis}
\end{tikzpicture}
\vspace{-0.1in}
\caption{Accuracy in uncertain user position coordinates}
\label{fig:ModelVsGurobi_Noise}
\vspace{-0.15in}
\end{figure}
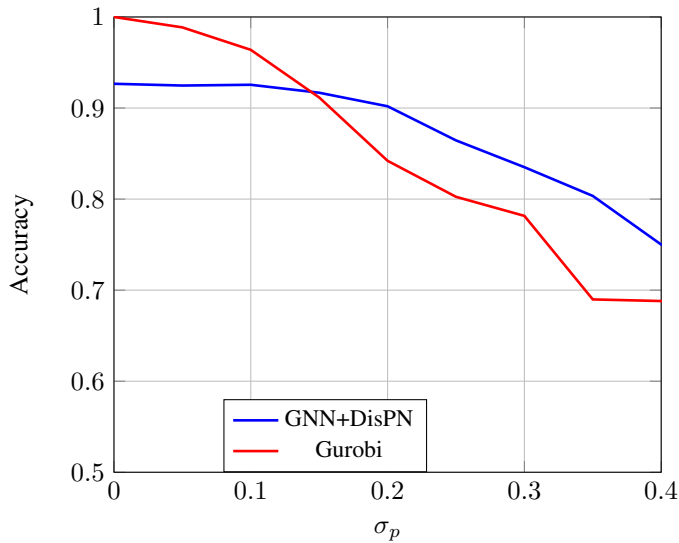

Figure~\ref{fig:ModelVsGurobi_Noise} provides insight into the robustness of our proposed approach under increasing levels of user position estimation errors. It compares the SNR accuracy of our model, using the Monte Carlo aggregation scheme described earlier, against the performance of a Gurobi-based optimizer that receives only a single noisy input realization. While Gurobi yields optimal solutions for a given input, its accuracy degrades significantly when run on imperfect estimates, as it lacks the ability to reason over uncertainty.
In contrast, our model maintains consistently higher SNR accuracy across all levels of standard deviation, $\sigma_p$ of position estimation errors, owing to its ability to aggregate predictions over multiple  samples. This highlights a key practical advantage: instead of relying on a single imprecise input, our model synthesizes a robust decision based on an ensemble of perturbed observations.

\begin{figure}[!t]
\centering
\begin{tikzpicture}
    \begin{axis}[
        width=1\linewidth,
        xlabel={$\sigma_p$},
        ylabel={Percentage of active antennas},
        ymin=0,
        ymax=0.5,
        xmin=0,
        xmax=1,
        grid=major,
        legend entries = {
            {GNN-DisPN},
        },
        legend style={font=\small, at={(0,0)}, anchor=south west},
    ]
    \addplot[blue, mark=x, line width=1pt]
    table {antenna_act_model.dat};
    \end{axis}
\end{tikzpicture}
\vspace{-0.1in}
\caption{Antennas activation with position estimation errors}
\label{fig:Activation_Noise}
\vspace{-0.15in}
\end{figure}
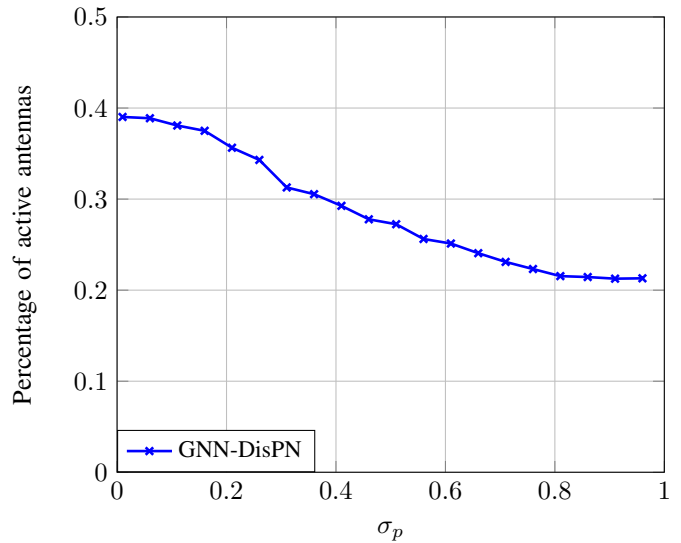

Fig.~\ref{fig:Activation_Noise} summarizes the model’s activation strategy under increasing levels of user position estimation errors. As the standard deviation, \(\sigma_p\), of the corresponding error increases, we observe a gradual reduction in the fraction of antennas activated by the model. This reflects a more conservative selection policy, which is beneficial in high-uncertainty scenarios: activating too many antennas under noisy inputs may lead to phase misalignments and destructive interference, degrading the overall SNR. Instead, by selectively activating fewer, more confident antennas, the model mitigates this risk and preserves signal quality. The results demonstrate that the augmented GNN+DisPN policy adapts to the reliability of the input, maintaining strong performance (from 91.5\% down to 82\% SNR accuracy) even when the user position is imprecisely known.

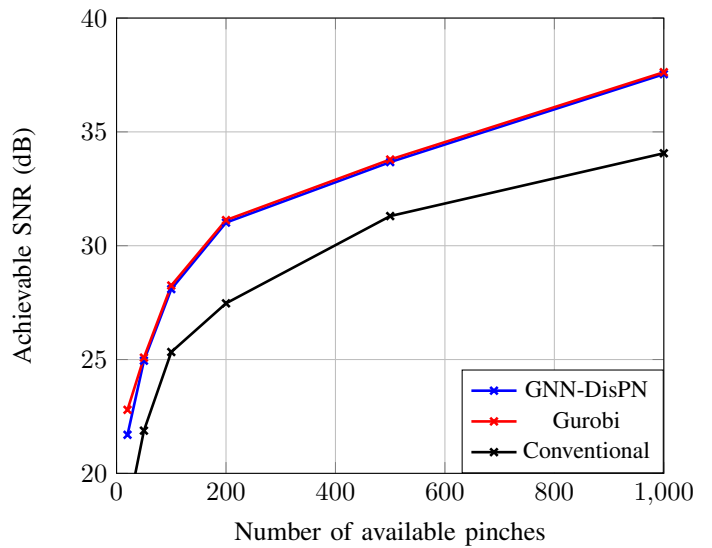
\begin{figure}[!t]
\centering
\begin{tikzpicture}
    \begin{axis}[
        width=1\linewidth,
        xlabel={Number of available pinches},
        ylabel={Achievable SNR (dB)},
        ymin=20,
        ymax=40,
        xmin=0,
        xmax=1000,
        grid=major,
        legend entries = {
            {GNN-DisPN},
            {Gurobi},   
            {Conventional},
        },
        legend style={font=\small, at={(1,0)}, anchor=south east},
    ]
    \addplot[blue, mark=x, line width=1pt]
    table {snr_number_of_pinches_model.dat};
    \addplot[red, mark=x, line width=1pt]
    table {snr_number_of_Pinches_gurobi.dat};    
    \addplot[black, mark=x, line width=1pt]
    table {snr_number_of_Pinches_conv.dat};
    \end{axis}
\end{tikzpicture}
\vspace{-0.1in}
\caption{Achievable SNR vs N available antennas.}
\label{fig:SNR_vs_N_active}
\vspace{-0.15in}
\end{figure}

Finally, Fig. \ref{fig:SNR_vs_N_active} illustrates the impact of the number of available pinches on the achievable SNR. To further demonstrate the ability of the proposed approach to mitigate path-loss, we compare its performance with that of an idealized conventional large-scale MIMO system. In this benchmark system, the number of antennas is equal to the number of available pinches, and all antennas are assumed to be co-located at the center of the area under consideration, at a height $H$, i.e., at the position $(0,0,H)$. In contrast to the proposed PAS, this conventional system assumes perfect constructive signal addition at the user via beamforming. Notably, the proposed PAS yields a gain exceeding 3 dB compared to the conventional system for a reasonable number of available PAs. Another important observation is that, despite its significantly lower computational complexity, the GNN+DisPN approach achieves nearly identical performance to Gurobi across all considered values of available pinches, thereby demonstrating the scalability of the proposed method.

\section{Conclusion}

In this work, we investigated the challenging problem of optimizing antenna activation in PASs in order to maximize the user's rate. We formulated the task as a QF01P problem and proposed three neural network architectures of increasing complexity to approximate optimal activation policies: a simple MLP with fusion, a GNN-enhanced model, and a distributed attention-based GNN+DisPN.
Through comprehensive experiments, we demonstrated the superiority of graph-based and attention-augmented models in capturing complex spatial dependencies and adapting to user-specific configurations. Notably, the GNN+DisPN model achieved state-of-the-art accuracy, generalizing effectively even to large-scale antenna arrays and showing resilience to user localization uncertainty.
These findings underscore the potential of deep learning, and especially graph-based reasoning, in efficiently solving combinatorial wireless resource allocation problems. 

\appendix
\subsection{Q Matrix Decomposition and Generalization}
\label{Appendix1}
In our specific setting, the matrix \( \mathbf{Q} \in \mathbb{R}^{n \times n} \) is constructed as
\begin{equation}
\mathbf{Q} = \Re(\mathbf{B} \mathbf{B}^H),
\end{equation}
where \( \mathbf{B} \in \mathbb{C}^{n \times 1} \) is a complex-valued antenna response vector. Letting \( \mathbf{B} = \mathbf{r} + j\mathbf{i} \), with \( \mathbf{r}, \mathbf{i} \in \mathbb{R}^n \), this expands as
\begin{equation}
\mathbf{Q} = \mathbf{r} \mathbf{r}^\top + \mathbf{i} \mathbf{i}^\top,
\end{equation}
revealing that \( \mathbf{Q} \) is a real, symmetric, positive semidefinite matrix of rank at most 2. Consequently, it admits an eigen-decomposition of the form:
\begin{equation}
\mathbf{Q} = \lambda_1 \mathbf{u}_1 \mathbf{u}_1^\top + \lambda_2 \mathbf{u}_2 \mathbf{u}_2^\top,
\end{equation}
with \( \lambda_1, \lambda_2 \geq 0 \) and \( \mathbf{u}_1, \mathbf{u}_2 \in \mathbb{R}^n \) orthonormal. In this case, the real and imaginary parts of \( \mathbf{B} \) lie in the subspace spanned by \( \mathbf{u}_1 \) and \( \mathbf{u}_2 \), and a compatible complex vector \( \mathbf{B} \) satisfying \( \mathbf{Q} = \Re(\mathbf{B} \mathbf{B}^H) \) can be written as:
\begin{equation}
\mathbf{B} = \alpha \mathbf{u}_1 + j \beta \mathbf{u}_2,
\end{equation}
for real scalars \( \alpha, \beta \) such that \( \alpha^2 = \lambda_1 \), \( \beta^2 = \lambda_2 \), up to a global complex phase. Although this reconstruction is not unique, it confirms that the structure of \( \mathbf{B} \) is encoded in the spectral components of \( \mathbf{Q} \).

In our setting, the \( \mathbf{B} \) values are directly available and serve dual purposes: they define the matrix \( \mathbf{Q} \) used in combinatorial optimization (e.g., via Gurobi), and they act as edge features in the deep learning models. In contrast, in general QF01P problems where only the matrix \( \mathbf{Q} \) is known, one could attempt to approximate surrogate \( \mathbf{B} \)-like values by decomposing \( \mathbf{Q} \), even when its rank exceeds 2. For instance, if \( \mathbf{Q} \) is of rank \( r \), it can be expressed as
\begin{equation}
\mathbf{Q} = \sum_{k=1}^r \lambda_k \mathbf{u}_k \mathbf{u}_k^\top,
\end{equation}
and a generalization of \( \mathbf{B} \) could be formed by combining these modes as complex-valued components:
\begin{equation}
\mathbf{B} = \sum_{k=1}^r \gamma_k \mathbf{u}_k,
\end{equation}
where \( \gamma_k \in \mathbb{C} \), and coefficients are chosen such that \( \Re(\mathbf{B} \mathbf{B}^H) \approx \mathbf{Q} \). While the exact inverse mapping is not unique or always feasible, this idea suggests a pathway to extend the proposed learning-based solution to broader classes of quadratic fractional binary optimization problems, using \( \mathbf{Q} \) as the starting point for constructing compatible feature representations.

\subsection{Dinkelbach's Algorithm}
\label{Appendix2}
The Dinkelbach's algorithm is an iterative method for solving fractional programming problems of the form
\begin{equation}
\max_{\mathbf{x} \in \mathcal{X}} \frac{f(\mathbf{x})}{g(\mathbf{x})},
\end{equation}
where $f$ and $g$ are continuous functions and $\mathcal{X}$ is the feasible set. In the considered optimization problem in \eqref{eq:our_qf01p}, the objective has the form
\begin{equation}
\max_{\mathbf{x} \in \{0,1\}^n} \frac{\mathbf{x}^T \mathbf{Q} \mathbf{x}}{\mathbf{c}^T \mathbf{x}},
\end{equation}
where $\mathbf{Q} \in \mathbb{R}^{n \times n}$ is a positive semi-definite matrix and $c \in \mathbb{R}^n$ is a positive cost vector.

The algorithm transforms the fractional objective into a sequence of easier subproblems. At iteration $k$, given $\lambda^{(k)}$, it solves
\begin{equation}
\mathbf{x}^{(k)} = \arg\max_{\mathbf{x} \in \{0,1\}^n} \left( \mathbf{x}^T \mathbf{Q} \mathbf{x} - \lambda^{(k)} \mathbf{c}^T \mathbf{x} \right),
\end{equation}
and updates
\begin{equation}
\lambda^{(k+1)} = \frac{(\mathbf{x}^{(k)})^T \mathbf{Q} \mathbf{x}^{(k)}}{\mathbf{c}^T \mathbf{x}^{(k)}}.
\end{equation}

The process repeats until convergence, typically when the change in $\lambda^{(k)}$ falls below a prescribed tolerance. In this way, the Dinkelbach's method reduces a difficult fractional problem to a sequence of quadratic maximizations with linear penalties.

The Dinkelbach's algorithm guarantees convergence to the global optimum if each subproblem is solved exactly. In our case, the fractional objective
\begin{equation}
\phi(x) = \frac{\mathbf{x}^T \mathbf{Q} \mathbf{x}}{\mathbf{c}^T \mathbf{x}}
\end{equation}
is a ratio of convex functions, and the feasible set $\{0,1\}^n$ leads to subproblems that are concave maximizations (since $Q$ is positive semi-definite). The key property is that $x^*$ is optimal if and only if
\begin{equation}
\max_{x \in \{0,1\}^n} \left( \mathbf{x}^T \mathbf{Q} \mathbf{x} - \lambda^* \mathbf{c}^T \mathbf{x} \right) = 0,
\end{equation}
where $\lambda^* = \phi(x^*)$. Each subproblem is solved optimally using the Gurobi optimizer with the branch and bound method. Thus, the  Dinkelbach’s method ensures that the final solution is globally optimal for the original PAs problem.

\vspace{-0.1in}

\bibliographystyle{IEEEtran}
\bibliography{bib}

\end{document}